\documentclass{article}
\usepackage{amssymb}
\usepackage{amsthm}
\usepackage{amsmath}
\usepackage{microtype}
\usepackage{graphicx}
\usepackage{subcaption}
\usepackage{booktabs} 

\usepackage{hyperref}

\usepackage[accepted]{icml2021}

\icmltitlerunning{RL Agent Training with Goals for Real World Tasks}

\begin{document}
\newtheorem{theorem}{Theorem}[section]
\newtheorem{definition}{Definition}[section]

\twocolumn[
\icmltitle{Reinforcement Learning Agent Training with Goals for Real World Tasks}




\begin{icmlauthorlist}
\icmlauthor{Xuan Zhao}{first}
\icmlauthor{Marcos Campos}{first}
\end{icmlauthorlist}

\icmlaffiliation{first}{Microsoft}

\icmlcorrespondingauthor{Xuan Zhao}{zhao.xuan@microsoft.com}
\icmlkeywords{Reinforcement Learning, Task Specification, TLTL, GOAL, BONSAI, ICML}

\vskip 0.3in
]



\printAffiliationsAndNotice{} 

\begin{abstract}
Reinforcement Learning (RL) is a promising approach for solving various control, optimization, and sequential decision making tasks. However, designing reward functions for complex tasks (e.g., with multiple objectives and safety constraints) can be challenging for most users and usually requires multiple expensive trials (reward function “hacking”). In this paper we propose a specification language (Inkling Goal Specification) for complex control and optimization tasks, which is very close to natural language and allows a practitioner to focus on problem specification instead of reward function hacking. The core elements of our framework are:  (i) mapping the high level language to a predicate temporal logic tailored to control and optimization tasks, (ii) a novel automaton-guided dense reward generation that can be used to drive RL algorithms, and (iii) a set of performance metrics to assess the behavior of the system. We include a set of experiments showing that the proposed method provides great ease of use to specify a wide range of real world tasks; and that the reward generated is able to drive the policy training to achieve the specified goal.
\end{abstract}

\section{Introduction}
A key shortcoming of RL is that the user must manually encode the task as a real-valued reward function, which can be challenging for several reasons: the need for a relatively strong math background and rich domain knowledge about the target environment. For complex problems, especially the ones involving temporal dependencies, a reward function may not be available. Furthermore, the user often needs to manually shape the reward to guide the agent towards learning the problem, however, reward shaping can be very error prone, and if done incorrectly, can shift the optimal policy from the one optimum for the original problem.

To address the ease of use and complexity of generating rewards in RL we propose a language for users to specify what they want to achieve. The proposed language allows the user to specify objectives using high level primitive operators over states, and then combine these primitive operators using logical or temporal operators. Our language is a higher level abstraction of a formal task specification language, which is an extension to Linear Temporal Logic (LTL). LTL or similar formal languages provide a logical way of expressing a task, however, it needs some learning curve to use. Our language is a near-natural language abstraction of this formal task specification language, making it generally accessible. The principle underlying our approach is that users just need to know what they want to achieve, no need to know how to make the agent achieve it. Overall, the user just provides the high-level task structure, and proposed framework instantiates the necessary low-level details.  

Under the hood, the higher level task specification will be translated to a formal task specification language ETLTL (Extended Truncated Linear Temporal Logic). ETLTL is an extension to TLTL (Truncated Linear Temporal Logic) in order to specify some very common optimization problems, like minimize a variable, drive an object to a certain region, etc. The ETLTL specification will then be translated to its automaton counterpart, Semi-Finite State Predicate Automaton (SFSPA), whose design allows us to solve common control and optimization problems. The proposed framework leverages the properties of SFSPA to generate a reward to drive a RL algorithm. The reward generated will be a dense reward, bypassing the need for doing manual reward shaping. 

We also proposed new metrics for assessing the behavior of the system. These metrics are very easy to interpret and directly reflect how the policy is doing towards satisfying the goal specification. 

To demonstrate its applicability and generality, we evaluated the performance of this approach on several real world problems. 

As a summary, our contributions are:
\begin{itemize}
\item  We proposed a near-natural specification language for specifying common control and optimization tasks.
\item We designed a formal task specification language called ETLTL, also designed an automaton called SFSPA, as the automaton counterpart of ETLTL. ETLTL together with SFSPA is able to specify and solve a large set of control and optimization problems found in industrial automation.
\item We designed an algorithm for compiling an ETLTL specification into a dense reward.
\item We introduced new assessment metrics for providing insights into the performance of the trained policy.
\end{itemize}

\section{Related Work}
In recent years there has been increasing interest in using Temporal Logic (TL) specifications as a formal and principle way for creating rewards for RL. The survey in \cite{survey} provides a good overview of these different approaches. One of the temporal logics language, called STL (Signal Temporal Logic) \cite{STL1} has attracted much attention due to the fact that it supports quantitative semantics, also known as robustness \cite{robustness} that can be used as reward. \cite{STL2} \cite{STL3} \cite{STL4} have successfully adopted STL robust semantics for translating STL specification to reward. \cite{robustness1} \cite{robustness2} \cite{robustness3} also improved upon the robustness score calculations proposed in \cite{robustness}, and got even better results. However, the reward generation process for STL is non-Markovian, and the reward is sparse as it can only be generated for a whole trajectory, making it very hard to solve complicated tasks. Also STL requires the designer to manually specify temporal bounds on how long each task should take, which is a problem for RL where length of execution can vary for different episodes.

TLTL \cite{TLTL} \cite{thesis} is another temporal logic language that supports robustness as quantitative semantics. TLTL formulas, unlike STL’s, can be evaluated against finite trajectories of any length, which makes it more suitable for RL applications. \cite{robotics} relies on TLTL and describes a predicate-based automaton, Finite State Predicate Automaton (FSPA), that works with TLTL to generate rewards. Automaton-based methods use an automaton to absorb the temporal dependency in the temporal logic, thus making the reward generation process Markovian. \cite{robotics} combines TLTL and FSPA as a formalism to specify finite-horizon, time-dependent (non-Markovian) robotic manipulation tasks over continuous state and action spaces, and generate the corresponding dense rewards. However, it's more targeted towards robotic manipulation tasks, so a trajectory just needs to get to the FSPA acceptance state, then the trajectory (an episode in RL setting) can terminate. It does not prescribe how to optimize the system further beyond the immediate arrival at the acceptance state.

The work presented here expands upon \cite{robotics}. 
In this work, we are targeting a wider range of problems where RL can be useful than \cite{robotics}, including optimization problems, e.g. minimize or maximize an objective function as far as possible subject to some constraints. For these optimization problems, we want to continue optimizing and improving the robustness of the whole system until the end of time limit.

\cite{NIPS} constructs a task monitor from the given specification, then use the task monitor to assign shaped rewards to the system. However, the construction of the task monitor can become very complicated when the task specification is a complex one. Although our work proposes a very different implementation for reward generation, it shares with \cite{NIPS} the use of a high level task specification language that is more readily accessible to a broader user base than any temporal language dialect.

\section{Preliminaries}
\subsection{Reinforcement Learning}
We follow \cite{sutton} for our version of the reinforcement learning problem where an agent interacts with an environment $E$ in discrete timesteps. At each timestep $t$, the agent observes a state $s_t \in \mathbb{R}^n$, performs an action $a_t \in \mathbb{R}^m$, transitions to a new state $s_{t+1} \in \mathbb{R}^n$, and receives feedback reward $r_t \in \mathbb{R}$ from the environment $E$. The goal of reinforcement learning is to optimize the agent’s action-selecting policy such that it achieves maximum expected reward.
The sequence $(s_0, a_0, ..., s_t, a_t, ..., s_T)$ is modeled as a Markov Decision Process (MDP) $\mathcal{M} = ⟨S, A, P, s_0, R_T, \gamma, T⟩$, where $S$ is a set of possible states, $A$ is the set of actions the agent is allowed to take, $P$ is the state transition probability defined as: $P: S\times A \times S \rightarrow [0, 1]$, $s_0$ is initial state, $R_T = \sum_{\tau=t}^T\gamma^{\tau - t}r_\tau(s_\tau, a_\tau)$, $\gamma \in (0, 1)$ is the discounted factor. $T$ is the episode termination time.

A control policy $\pi$ for the agent is a function $\pi : S \rightarrow A$ deterministically, or $\pi : S \times A \rightarrow [0, 1]$ stochastically. An RL algorithm is an optimization algorithm that can find the control policy $\pi(a_t|s_t)$ in order to maximize the reward return $R_T$ from the environment.

Trajectory or episode is a sequence of state and actions under a certain policy $\pi$: $\tau = (s_0, a_0, s_1, ..., s_t, a_t, s_{t+1}, a_{t+1}, ..., s_T)$ . In the case of a stochastic policy $\pi$, $p^{\pi}(\tau)$ represents the trajectory distribution \cite{survey}. For this paper, we are only considering finite trajectories, so $T < \infty$.
\subsection{Temporal Logics}
Temporal Logics (TLs) \cite{wikiTL, TL1, TL2, TL3, TL4} is any system of rules and symbolism for representing and reasoning about propositions qualified in terms of time. TLs provide an interface for specifying the task. We are looking for a TL specification that has qualitative and quantitative semantics. STL and TLTL both satisfy the needs. But the quantitative semantics generating process for STL is not Markovian. So we chose TLTL as a base temporal logics, and will modify it to serve a broader range of real world problems.

\subsection{Problem Formulation}\label{problemformulation}
Consider a continuous-time dynamical system as:
\begin{equation}\label{problem1}
    s(t) = f(s, a), s(0) = s_0
\end{equation}
where $t \in \mathbb{R}_{ 0\le t \le T}$, $T ⟨ \infty$, $s(t) \in S \subset \mathbb{R}^n$ is the state, $a(t) \in A \subset \mathbb{R}^m$ is the control input or action at time $t$, $s_0 \in S$ is the initial state. $\mathcal{T} = (s_0, a_0, s_1, ..., s_t, a_t, s_{t+1}, a_{t+1}, ..., s_T)$ is a trajectory.\\
{\itshape Problem:} Given a system denoted in equation (\ref{problem1}) and a Temporal Logic formula $\phi_f$ over predicates over state $s$, find a control input $a(t)$ such that the resulting trajectory $\mathcal{T}$ satisfies $\phi_f$ and maximizes quantitative semantics for $\phi_f$. 

Concretely, there are two types of problems we want to solve. One is control manipulation, where we want the agent to achieve a task, e.g. get ball then drop the ball to place $A$. As soon as the task is achieved, so $\phi_f$ is satisfied, you call it an end, and the trajectory can terminate. Another type of problem is optimization under constraint, e.g. maximize profit while ensuring the quality of products. For this type of problem, we want profit to be as high as possible, we can set a boundary by saying that it's acceptable if the boundary is reached, but the trajectory should not terminate at that point, and the higher we can push the profit, the better. Higher profit values should be encouraged with more reward.

$\phi_f$ won't be the direct task specification interface for our system due to its inaccessibility to general public. We need another user-friendly, near-natural language $\mathcal{N}_f$ to serve as interface.

\section{Approach}
In this section, we'll illustrate our approach for solving the problem. In section ~\ref{highspecification}, we'll discuss a near-natural language for task specification, called inkling goal specification; Section ~\ref{ETLTL} describes the proposed linear temporal logic dialect ETLTL and how to translate the inkling goal specification to ETLTL formulation; Section ~\ref{robustnessCalculation} discusses how to calculate the quantitative semantics - robustness for ETLTL; section ~\ref{SFSPA} introduces our modified version of Automaton - SFSPA, and proves SFSPA is the automaton counterpart of ETLTL; section ~\ref{rewardGeneration} describes the reward generation algorithm and also provides some practical tips towards the successful and effective application of the algorithm to real world problems.

\subsection{Inkling Goal Specification}{\label {highspecification}}

LTL or similar formal languages provide a logical way of expressing task and also provides great expressiveness, however, it involves quite a bit learning in order to use it correctly, so still inaccessible to general public or domain experts. We create a language that is a quasi natural language abstraction of this formal task specification language, making it generally accessible. This language is integrated into the Inkling language \cite{inkling} as Inkling Goal Specification.

In the current version, it supports the following operators:\\
\{reach \textbar drive \textbar avoid \textbar minimize \textbar maximize\}. \\
We'll also support \{then, until, and, or\} in future versions.\\
The syntax and meaning of the five supported operators are as follows:
\begin{itemize}
    \item {\itshape \small reach objectiveName: testValue in targetRange } -
    {\small Get to a target range as quickly as possible.}
    \item {\itshape \small drive objectiveName: testValue in targetRange } -
    {\small Get to a target range as quickly as possible and stay in range as much as possible.}
    \item {\itshape \small maximize objectiveName: testValue in targetRange } -
    {\small Push the test value as high as possible.}
    \item {\itshape \small minimize objectiveName: testValue in targetRange } - {\small Push the test value as low as possible.}
    \item {\itshape \small avoid objectiveName: testValue in avoidRange } - {\small Avoid the avoidRange.}
\end{itemize}

Note that both targetRange and avoidRange can be open ended with only one bound (e.g.: Goal.RangeAbove(100)) or closed range with two bounds (e.g.: Goal.Range(2, 50)). 

Users can use this language to specify what they want to achieve, without knowing how to achieve it (the underlying reward). Due to the richness this language provides, the specification of complicated tasks becomes possible. With this language being easily expressible, it greatly lowers the barrier for specifying and training an RL model for solving a specific problem.

\subsection{ETLTL}\label{ETLTL}
The inkling goal specification language described in \ref{highspecification} will be translated under the hood to a formal language specification. We designed a formal language ETLTL by extending TLTL \cite{thesis}. ETLTL should satisfy a large range of the specification needs from real world use cases of RL.

The Syntax is as follows:
\begin{equation}
\phi := \\
T | f(s) ⟨ c | \neg \phi | \phi \wedge \psi | 
\phi \vee \psi | \mathcal{F} \phi | \phi \mathcal{U} \psi | \mathcal{X} \psi  | G_{k} \psi
\end{equation}
where $T$ is the True Boolean constant. $s \in S$ is an MDP state.
$f(s) ⟨ c$ is a predicate over the MDP states where $c \in \mathbb{R}$, $f$ is a scalar function defined over the state space of the system; $\neg$ (negation/not) and
$\wedge$ (conjunction/and), $\vee$ (disjunction/or) are Boolean connectives. $\mathcal{F}$ (eventually),
$\mathcal{U}$ (until), $\mathcal{X}$ (next), are temporal operators \cite{thesis}.
ETLTL (Extended Truncated Linear Temporal Logic) is an extension to TLTL by including an operator  - time bounded always $G_k \psi$. $G_k \psi$ means that $\psi$ needs to hold for $k$ steps, where $k$ is often equal to the number of steps left in the episode, meaning that $\psi$ needs to hold until the end of the episode. With the addition of $G_k \psi$, it makes it possible to solve problems which need further improvement beyond the acceptance point, thus making more problems, like optimization problems, expressible. The operators we want to cover in the inkling goal specification will be translated to ETLTL as follows:
\begin{itemize}
    \item {\small \itshape minimize objectiveName: testValue in Goal.RangeBelow( upperbound ) $\rightarrow$ {\bf $G_k$($\mathcal{F}$(minimize testValue below upperbound))}}
    \item  {\small \itshape maximize objectiveName: testValue in Goal.RangeAbove( lowerbound ) $\rightarrow$ {\bf
    $G_k$($\mathcal{F}$(maximize testValue above lowerbound))}}
    \item {\small \itshape reach objectiveName: testValue in targetRange $\rightarrow$ {\bf   $\mathcal{F}$(testValue in targetRange)}}
    
    \item {\small \itshape drive objectiveName: testValue in targetRange $\rightarrow$ {\bf
    $G_k$($\mathcal{F}$(testValue in targetRange))}}
    \item  {\small \itshape avoid objectiveName: testValue in avoidRange $\rightarrow$ {\bf
    $G_k$($\neg$(testValue in avoidRange))}}
    \item {\small \itshape goalA AND goalB $\rightarrow$ {goalA $\wedge$ goalB}}
    \item {\small \itshape goalA OR goalB $\rightarrow$ {goalA $\vee$ goalB}}
    \item {\small \itshape goalA THEN goalB $\rightarrow$ $\mathcal{F}$(goalA $\wedge$ $\mathcal{X}$($\mathcal{F}$ (goalB)))}
    \item {\small \itshape  goalB UNTIL goalA $\rightarrow$  goalB $\mathcal{U}$ goalA}
    
\end{itemize}

\subsection{Robustness Calculation}\label{robustnessCalculation}
\label{sec:length}

Temporal languages can support two types of semantics:
\begin{itemize}
\item Qualitative semantics: A formula is either True or False;
\item Quantitative semantics: A continuous measure of formula satisfaction.
\end{itemize}
TLTL and STL both use robustness for quantitative semantics, robustness can also serve as reward that maps to the task specification. Modern control or optimization systems must satisfy complicated control objectives while withstanding a wide range of disturbances. The system being robust means that the system can still satisfy these objectives disregarding  these disturbances. Formally, if the system satisfies the formula with robustness $r$, then any disturbance of size less than $r$ cannot cause it to violate the formula. Robustness serving as reward can also provide a direct interpretation of the system, since the more reward the agent gets, the more robust is the trajectory towards satisfying the goal specification.

In this work, we use distance based robustness, so the robustness for variable $x$ approaching a region $A$ will be:
\begin{equation}
\rho(x \rightarrow A) = \lVert x - A_{center} \rVert
\end{equation}
where $A_{center}$ is the center of region A.

For Minimize/Maximize, the robustness is defined as the difference between the variable value and the target bound $c$, so the robustness will be:
\begin{itemize}
    \item For Minimize:
    $\rho(x ⟨ c) = c - x$
    \item For Maximize:
    $\rho(x ⟩ c) = x - c$
\end{itemize}

For logical expressions, we'll calculate robustness following the rules below \cite{robustness}:
\begin{itemize} \label{robustnesscal}
    \item $\rho (\phi \wedge \psi)$ = $\min{(\rho_\phi, \rho_\psi)}$
    \item $\rho (\phi \vee \psi)$ = $\max{(\rho_\phi, \rho_\psi)}$
    \item $\rho (\neg \phi)$ = $-\rho_\phi$
\end{itemize}

Note that the robustness scores calculated with $\min$ ($\wedge$) and $\max$ ($\vee$) are non-smooth and non-convex, suffers from masking effect and locality problem, also will lead to maximum system robustness being capped by the least robust subgoal, recent work \cite{robustness1} \cite{robustness2} has proposed robustness calculation method to replace the $\min$/$\max$. It is straightforward to replace the robustness calculation depicted in this section with the calculation proposed in \cite{robustness1} and \cite{robustness2}. We'll replace and experiment with these more advanced robustness calculation methods in future work. 

For temporal expressions, as all the temporal dependencies are already absorbed in the automaton transitions, we won't have predicates that have temporal dependencies on the automaton edge, we'll also see in the next section that only robustness of predicates on the edge needs to be calculated, thus we don't need to consider temporal dependencies when calculating robustness. 

Distance-based robustness provides a metric over the whole space for continuous variables. It can be used to create non-sparse rewards thus avoiding the need for reward shaping.

Robustness (and thus a reward based on it) can continue to grow even though Goal Satisfaction or Goal success is at $100\%$ because the system is finding more robust solutions. In optimization problems, like Minimize/Maximize problems, we want to continue pushing this robustness to be as high as possible. Reporting robustness as reward is insightful, meaningful, and interpretable.

\subsection{Semi Finite State Predicate Automaton (SFSPA)}
\label{SFSPA}
Given an ETLTL task specification, an automaton can be created to capture the required sequence of motions needed in order to successfully satisfy the task specification. 

The behavior of the system and the state of the automaton will be analyzed for each time step. A reward is generated based on the system state $s_{MDP}$ and the automaton state $s_{Automaton}$, i.e. $r(s, a)$, where $s = s_{MDP} \cup s_{Automaton}$. Since the transitions in Automaton are able to absorb the temporal dependencies, this reward generation process is Markovian, making harder problems, especially problems involving temporal dependencies feasible. 

Among the existing Automaton types that can represent a linear temporal logic, FSPA \cite{thesis} has most of our desired properties, as its transitions of states depend on the robustness of the predicates on the edge, instead of the truth values, so each transition can naturally lead to a robustness gain that can be used as step reward, and we can derive a reward generation algorithm following this property. However, FSPA was designed mainly for the robotic manipulation tasks, will accept and terminate a trajectory once it gets in an acceptance state, which means that the robot has completed the task. However, for problems like optimization problems, we still need to continue improving the robustness of the system until the end of allowed time period. For these problems, we shouldn't terminate the episode even when the acceptance state is hit, we should encourage the agent to stay in acceptance state and continue improving the robustness of the whole system.
With this in mind, we designed SFSPA, with the following definition:\\
\begin{definition} {SFSPA: }
 An SFSPA is a tuple $\mathcal{A} = ⟨\mathcal{Q}, \mathcal{S}, \mathcal{E}, \mathcal{P}_e, q_0, F, Tr⟩$, 

where $\mathcal{Q}$ is a finite set of automaton states; $\mathcal{S} \subset \mathbb{R}^n$ is the MDP state; $\mathcal{E} \subset \mathcal{Q} \times \mathcal{Q}$ is the set of transitions in the automaton; $\mathcal{P}_e$ is the set of boolean formula predicates attached to the transition edge $e \in \mathcal{E}$. $q_0 \in \mathcal{Q}$ is the initial automaton state. $F\subseteq \mathcal{Q}$ is the final or acceptance state, $Tr \subset \mathcal{Q} $ is the trap state.

Except for trap state $Tr$, all other states can make transition back to itself.
\end{definition}
The edge allowing a state to transit back to itself is a Self-Loop Edge as defined below:
\begin{definition}{Self-Loop Edge: }
Edge that leads from the current automaton state $q$ back to $q$. $e_{q\rightarrow q} := q \rightarrow q$, where $q \in \mathcal{Q}$, $e_{q\rightarrow q}\in \mathcal{E}$.

\end{definition}
The main difference from FSPA lies in the acceptance state.

FSPA does not have a self-loop edge at the acceptance state, and will accept and terminate a trajectory once it reaches the acceptance state; while SFSPA allows a self-loop edge at the acceptance state, and will only accept a trajectory when it reaches an acceptance state with no edge out. If the acceptance state still has an edge out, the trajectory will not be accepted nor terminated, it will continue going until the time limit is up. During this time, agent will learn to take actions trying to make the system more robust regarding to the target formula, thus maximize the reward gain. 

At acceptance state, we'll use the robustness of the predicate on the self-loop edge as reward, encouraging the system to stay in the acceptance state, or visit the acceptance state as many times as possible.  This design allows the acceptance state to be revisited more than once. As long as at the end of the episode, the automaton is still in the acceptance state, the trajectory will be accepted. 

\begin{theorem}\label{theorem}
Given an ETLTL formula $\phi$ over predicates in a state set $\mathcal{S}$, for $\forall \mathcal{T}_{t=0..t}$ that satisfy $\phi$, there exists an SFSPA $\mathcal{A_\phi} = ⟨\mathcal{Q_\phi}, \mathcal{S}, \mathcal{E_\phi}, \mathcal{P}_{\phi, e}, q_{\phi, 0}, F_\phi, Tr_\phi⟩$ that accepts $\mathcal{T}_{t=0..t}$,
where $\mathcal{T}_{t=0..t}$ is a trajectory over $\mathcal{S}$ that spans over time steps $[0.. t]$.
\end{theorem}

\begin{proof}
We can prove this theorem using induction. Say the ETLTL formula $\phi$ has $n$ operators. 

{\bf Base case}: When $n = 1$, if the operator is not $G_k$, then according to \cite{robotics}, there exists an FSPA that accepts $\mathcal{T}_{t=0..t}$. If the operator is $G_k$, the ETLTL formula will be $G_k(\psi)$, where $\psi$ is a TLTL formula, $T$ or $f(s) < c$. The automaton to represent $\psi$ should be only two states, initial state, and when $\psi$ is true, it goes to acceptance state. Once it goes to acceptance state, we just need to stay in the acceptance state $q_f$ by adding and going through the self-loop edge $q_f \rightarrow q_f$ until the episode time limit, thus creating an SFSPA that accepts $\mathcal{T}_{t=0..t}$.

{\bf Induction:} 
Assume when $n = N-1$, theorem \ref{theorem} holds, we want to prove that theorem \ref{theorem} also holds for $n=N$, then by the principle of induction, theorem \ref{theorem} will hold for all $n \in \mathbb{Z}_+$.

So ETLTL formula $\phi_{N-1}$ of $N-1$ operators satisfy theorem \ref{theorem}, then there exists an SFSPA $\mathcal{F}_{N-1}$ that represents $\phi_{N-1}$. If the additional operator $\phi_{N}$ has is a TLTL operator, then according to \cite{robotics}, there will also be an SFSPA representing $\phi_N$. If the additional operator is $G_k$, $\phi_N = G_k(\phi_{N-1})$. Based on the definition of $G_k$, if $\mathcal{T}_{t=0..t}$ satisfies $G_k(\phi_{N-1})$, there will be a subtrajectory $\mathcal{T}'_{t=0..t'}$, where $t' <= t$ that satisfies $\phi_{N-1}$. So at $t = t'$, $\mathcal{F}_{N-1}$ is at acceptance state $q_f$. If $\phi_{N-1}$ doesn't have temporal operator $\mathcal{F}$, it can just stay in that acceptance state by adding and going through the self-loop edge $q_f \rightarrow q_f$ until the episode time limit; this result in an SFSPA that accepts $\mathcal{T}_{t=0..t}$. If $\phi_{N-1}$ has the temporal operator $\mathcal{F}$, the trajectory can choose to leave and go back to the acceptance state through the same automaton states that first lead to the acceptance state $q_f$. So we still have an SFSPA can accept $\mathcal{T}_{t=0..t}$.

Now through the principle of induction, we know that theorem \ref{theorem} will hold for $\forall n \in \mathbb{Z}_+$. Theorem \ref{theorem} is thus proved.
 
\end{proof}

From this theorem we know that an ETLTL formula can translate to an SFSPA that is able to represent the ETLTL formula. With the ETLTL and SFSPA, it is possible to solve optimization problems like Maximize/Minimize, drive as close as possible to a point, etc.

Next section, we'll discuss our reward generation algorithm, and how we take advantage of SFSPA to generate reward for encouraging optimizing a system as far as possible.

\subsection{Reward Generation Algorithm}\label{rewardGeneration}
As communicated in section \ref{robustnessCalculation}, ETLTL can use robustness as its quantitative semantics, which can be used as reward.

To derive reward from the automaton, we have the algorithm as stated in Algorithm ~\ref{algorithm}.

\begin{algorithm}[tb]
\caption{Algorithm for Reward Generation}
\begin{algorithmic}
    \STATE Initialize MDP state as $s_{MDP}^0$
    \STATE Initialize Automata state as $s_{Automata}^0$
    \STATE Initialize reward as 0
    \WHILE {$t ⟨ t_{max}$}{
 \STATE Action $\leftarrow$ policy($s_{MDP}^t$, $s_{Automata}^t$)
 \STATE Step Action in MDP state $\rightarrow$ $s_{MDP}^{t+1}$
 \STATE Use $s_{MDP}^{t+1}$ to step in Automata
	\IF {any of the outgoing edges has a True predicate}{ 
		\STATE transit Automata using that edge $e$: $s_{Automata}^t \rightarrow s_{Automata}^{t+1}$.
		
		\STATE reward += $\rho(p_e)$ ($p_e$ is the predicate attached to edge $e$)
	}
	\ELSE { 
		\STATE stay in the current Automata state
		\STATE reward += $\max_{e \in allOutgoingEdges}(\rho(p_e))$
    }
    \ENDIF
    
    \IF  {$s_{Automata}^t$ is acceptance state and There is an outgoing edge}{
		\STATE reward += $\rho(p_{selfLoopEdge})$ 
		}
	\ELSE {
		\STATE reward += $\max_{e \in allOutgoingEdges}(\rho(p_e))$
	}
    \ENDIF

	\IF {$s_{Automata}^{t+1}$ is acceptance state and there is no outgoing edge or $s_{Automata}^{t+1}$  is trap state}{
		\STATE terminate the episode
		\STATE reward += terminalReward
		\STATE break loop
    } 
    \ENDIF
    
\STATE $s_{MDP}^t$ $\leftarrow$ $s_{MDP}^{t+1}$
\STATE $s_{Automata}^t$ $\leftarrow$ $s_{Automata}^{t+1}$
\STATE $t = t + 1$
}

\ENDWHILE
\STATE return reward
\label{algorithm}
\end{algorithmic}
\end{algorithm}

{\bf Terminal Condition:} The episode terminates if any of the following conditions is true:
\begin{itemize}
\item Reaches acceptance state that has no outgoing edge
\item Reaches a Trap state (cannot exit) 
\item Reaches the maximum number of episode steps
\end{itemize}

the difference of this algorithm from the one depicted in \cite{robotics} \cite{thesis}, is when we arrive in the acceptance state, if there is still an outgoing edge from the acceptance state, we won't terminate the episode immediately. We'll use the robustness of the predicate on the self-loop edge as step reward. Then if the robustness of the self-loop edge is positive, the agent stays in the state and gains a positive step reward, thus encouraging the agent to stay in the state. This gives the agent a chance to further optimize the policy and further improve the system robustness even after the goals are achieved.

Some practical tips we need to apply for making this algorithm applicable to real world problems are: the scaling/normalization/boosting of robustness, the capping of step reward, and add terminal reward for early termination. Please refer to Appendix \ref{tips} for a detailed illustration on how the above are performed.

\subsection{Goal Assessment Metrics}
After a policy is trained, it needs to be tested to assess its viability. When we use a reward function directly, only a real value of reward can be shown to reflect the performance of the policy, however, reward often times is not meaningful, one needs to know the typical and best reward value for a certain problem to know how a specific policy is performing. 

With the goals specified, we can measure how likely we can or can not reach the goal, if we did not reach the goal, how far away we are, and what’s the possible reason of failing. This is a much richer feedback than a single reward value when we only have a reward function provided. With this richer feedback, it's much more direct to find out how the policy is performing against achieving the specified goal, and it's easier to figure out how to refine the training and adjust the goal specification, etc.

The goal assessment metrics we are currently reporting and how they are calculated are detailed in Appendix \ref{goalassessmentmetrics}.

\section{Experiments}
We applied the proposed approach to several real world problems, including: Ball balancing using the Moab robot \cite{moab, moab1}, factory logistics optimization \cite{factorylogistics}, tank filling \cite{tankfill}, chemical reactor optimization \cite{chemicalprocessing} and a drone control problem. From these experiments, we show that the reward generation process proposed in this paper is able to drive the policy training for a wide range of real world problems, and can guide the agent to successfully achieve the goal that was specified.

\begin{figure}
    \centering
    \begin{subfigure}[b]{\columnwidth}
        \includegraphics[width=\columnwidth]{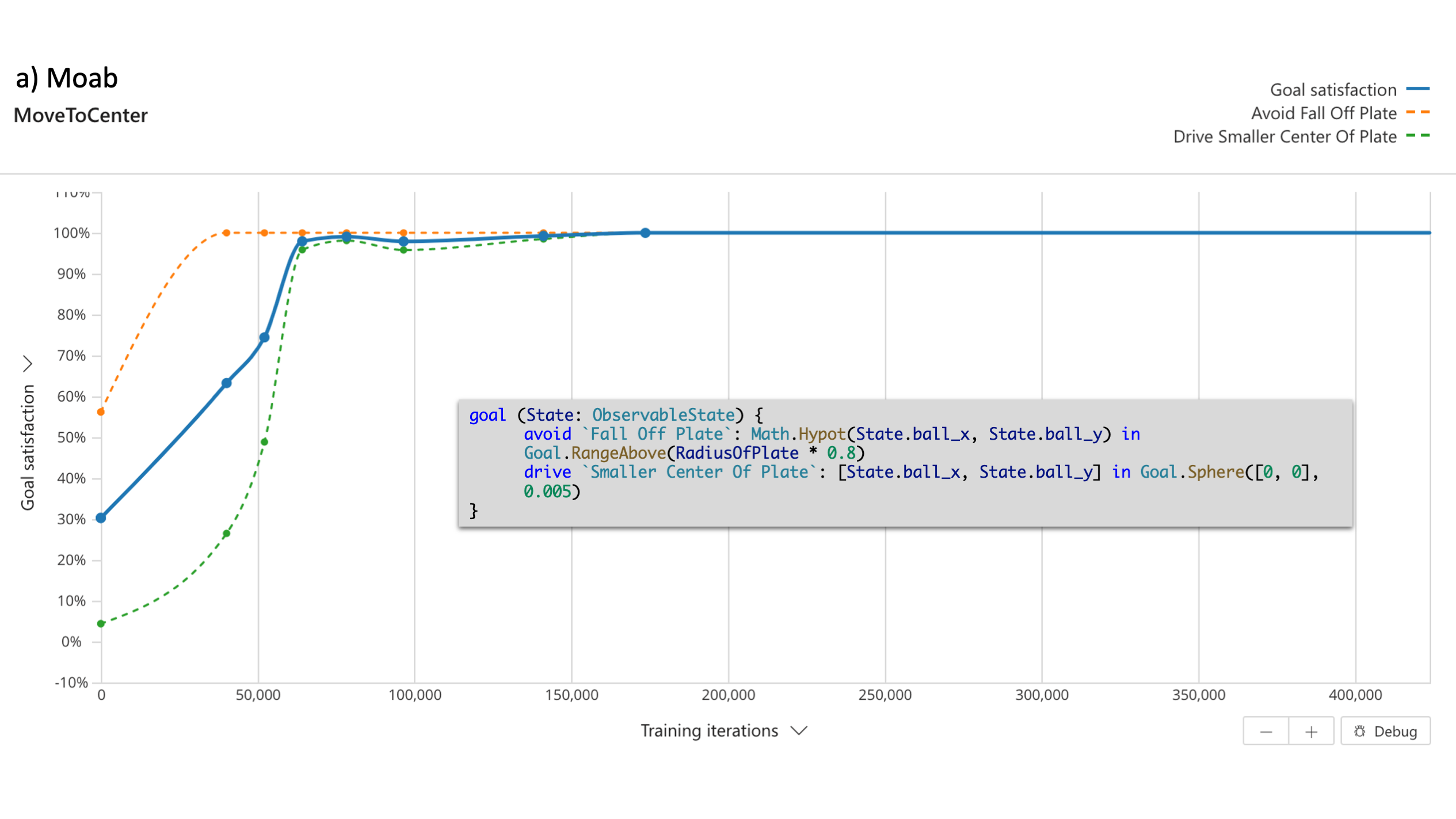}
        \caption{Balance ball to a small target region with radius 0.005m, while avoiding falling off the plate on the Moab device; RadiusOfPlate = 0.1125m}
        \label{moabp}
    \end{subfigure}
    \begin{subfigure}[b]{\columnwidth}
        \includegraphics[width=\columnwidth]{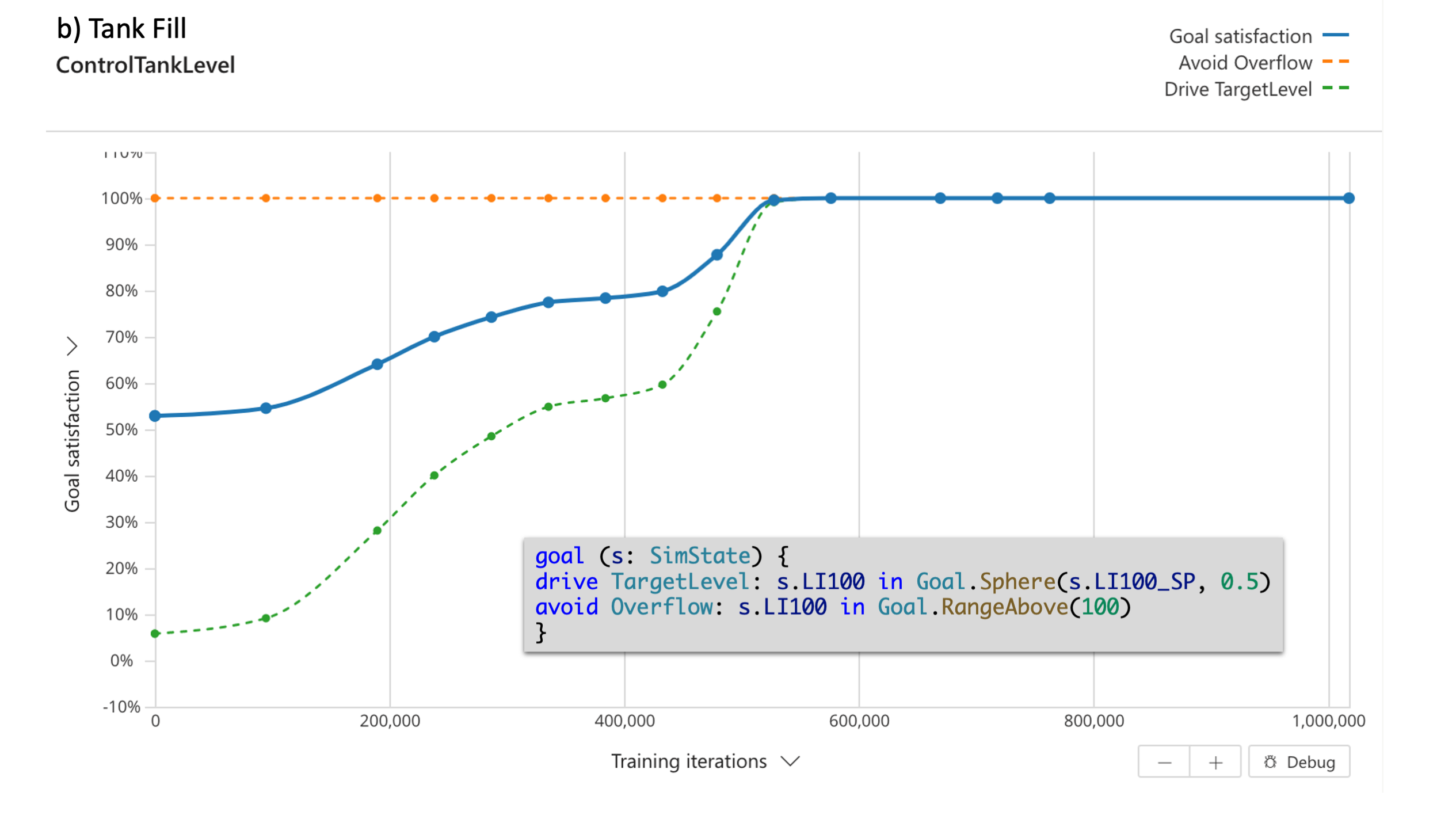}
        \caption{Control the filling of a tank to a certain level while avoiding tank overflowing}
        \label{tankfillp}
    \end{subfigure}
    \begin{subfigure}[b]{\columnwidth}
        \includegraphics[width=\columnwidth]{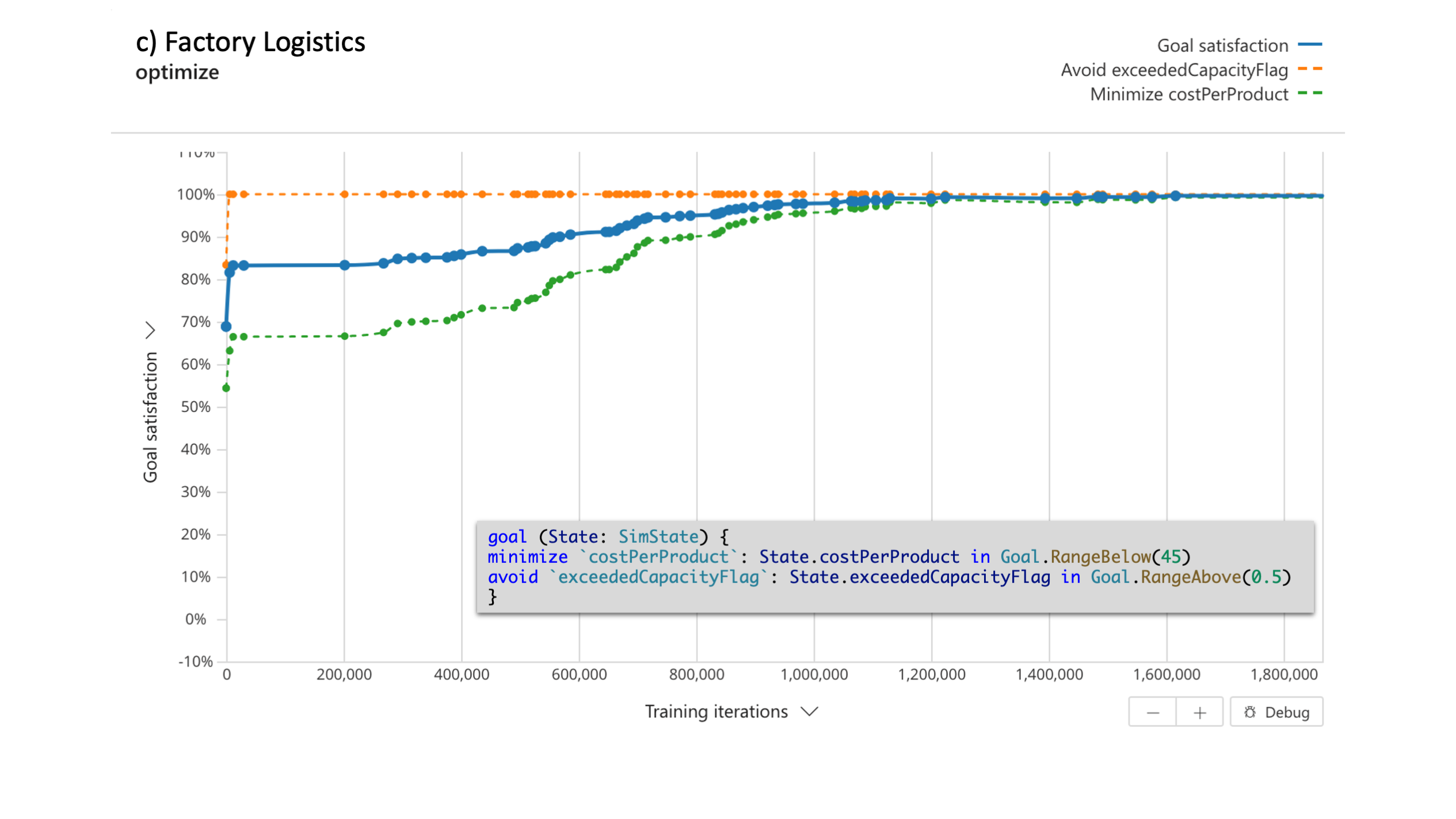}
        \caption{Minimize the operational cost of product manufacturing while limiting the resource usage to a certain level}
        \label{factorylogisticsp}
    \end{subfigure}

    \begin{subfigure}[b]{\columnwidth}
        \includegraphics[width=\columnwidth]{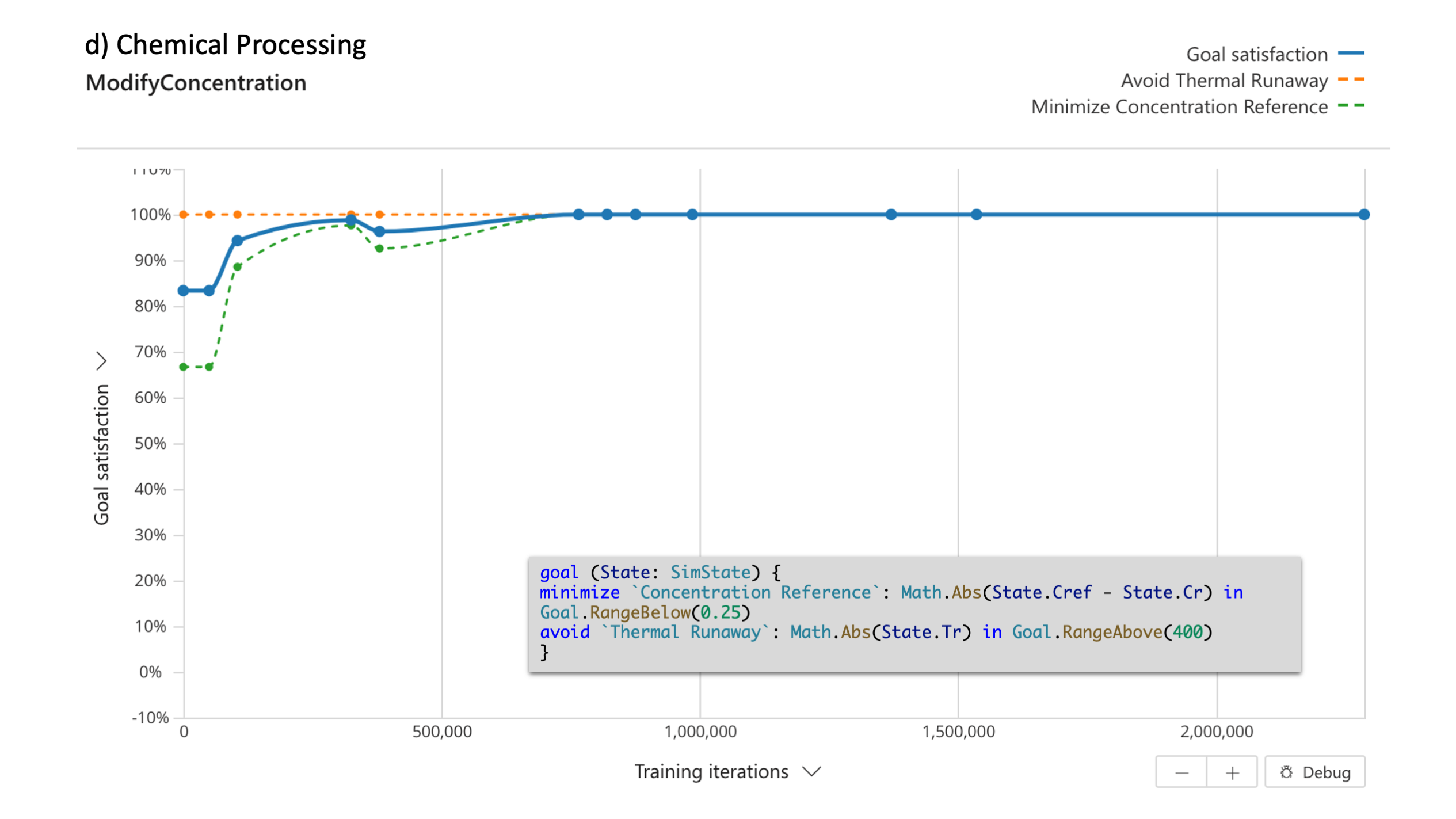}
        \caption{Drive residual concentration close enough to the reference concentration point under the constraint that temperature should not go beyond limit during this process}
        \label{chemicalprocessingp}
    \end{subfigure}
    \caption{Training policies with goals for real world problems. Plot is showing goal satisfaction rate, whose meaning and calculation formula are as explained in Appendix \ref{goalassessmentmetrics}}

\end{figure}
The objective of the ball balancing task using Moab is to teach the robot to balance a ball in the center of a plate. We used a DRIVE goal with a small target region, together with an AVOID goal, the goals used are in Figure \ref{moabp}. The robot is able to accomplish the task pushing the ball to a very small region of radius 0.005m (Figure \ref{moabp}). 

For the tank filling task (detailed setup in \cite{tankfill}) the objective is to fill a tank to any set point level between 5$\%$ to 95$\%$ without overflowing the tank or letting it get empty. The system is able to drive the tank level to the desired level successfully (Figure \ref{tankfillp}). 

The factory logistics optimization problem optimizes factory logistics for productivity and effective resource allocation while minimizing the operational cost of product manufacturing (detailed setup in \cite{factorylogistics}). Figure \ref{factorylogisticsp} shows the results for the task. With the goal specification provided the system learned to minimize the cost per product keeping it under a predefined maximum cost boundary while not exceeding the capacity of available resources. 

In the chemical reactor optimization task, the process considered here is a Continuous Stirred Tank Reactor (CSTR) during transition from low to high conversion rate (high to low residual concentration). Because the chemical reaction is exothermic (produces heat), the reactor temperature must be controlled to prevent a thermal runaway. The control task is complicated by the fact that the process dynamics are nonlinear and transition from stable to unstable and back to stable as the conversion rate increases. The controlled variables (states) are the residual concentration and the reactor temperature, and the manipulated variable (action) is the temperature of the coolant circulating in the reactor's cooling jacket (detailed setup in \cite{chemicalprocessing}). Using the approach in this paper the system is able to successfully drive residual concentration close enough to the reference concentration point under the constraint that temperature should not go beyond limit during this process (Figure \ref{chemicalprocessingp}).

\begin{figure}
    \centering
    \begin{subfigure}[b]{\columnwidth}
        \includegraphics[width=\columnwidth]{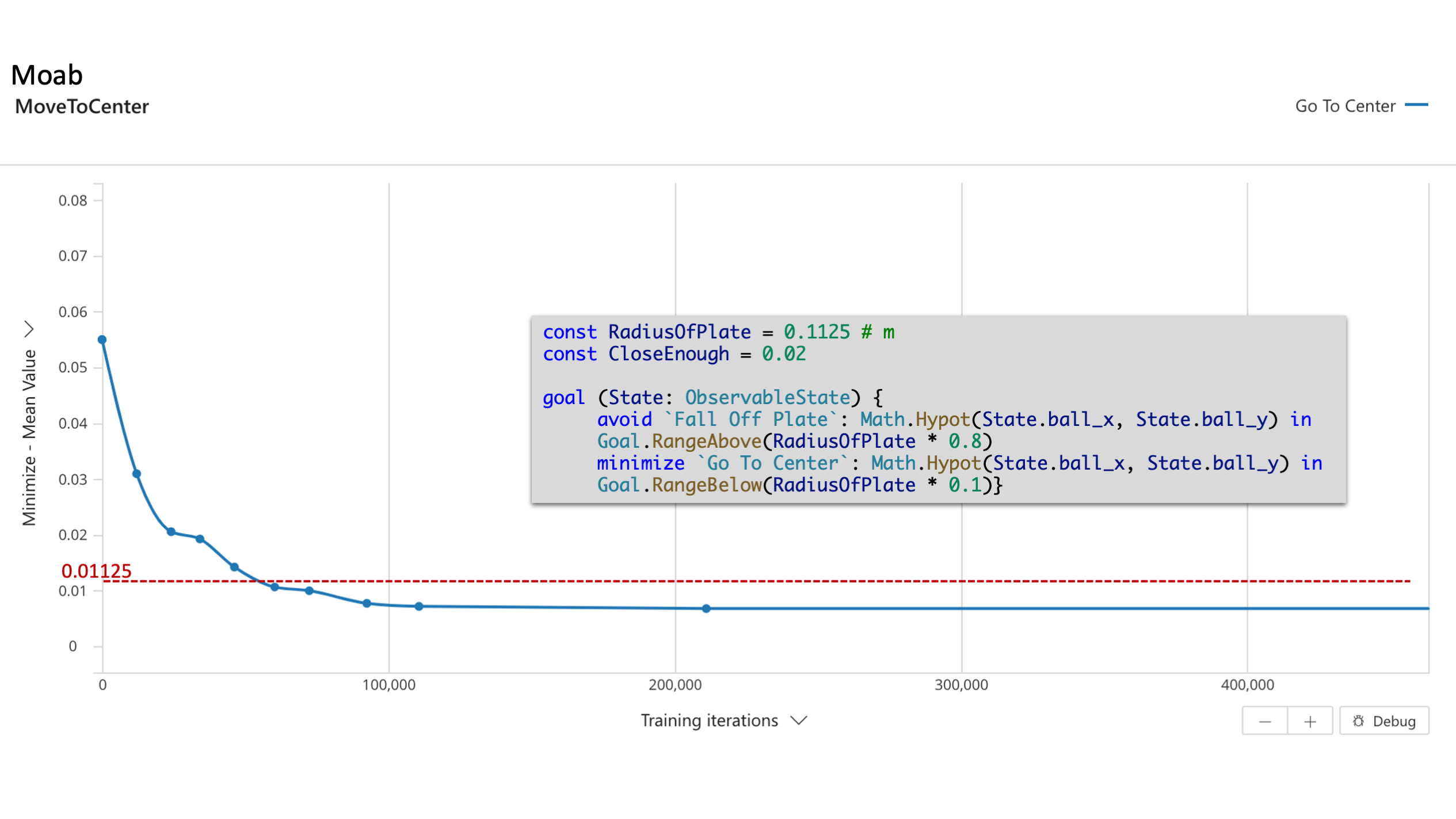}
        \caption{Minimize the distance from ball to plate center to below $0.01125$ on the Moab Device \cite{moab, moab1}}
        \label{minimize}
    \end{subfigure}
    \begin{subfigure}[b]{\columnwidth}
        \includegraphics[width=\columnwidth]{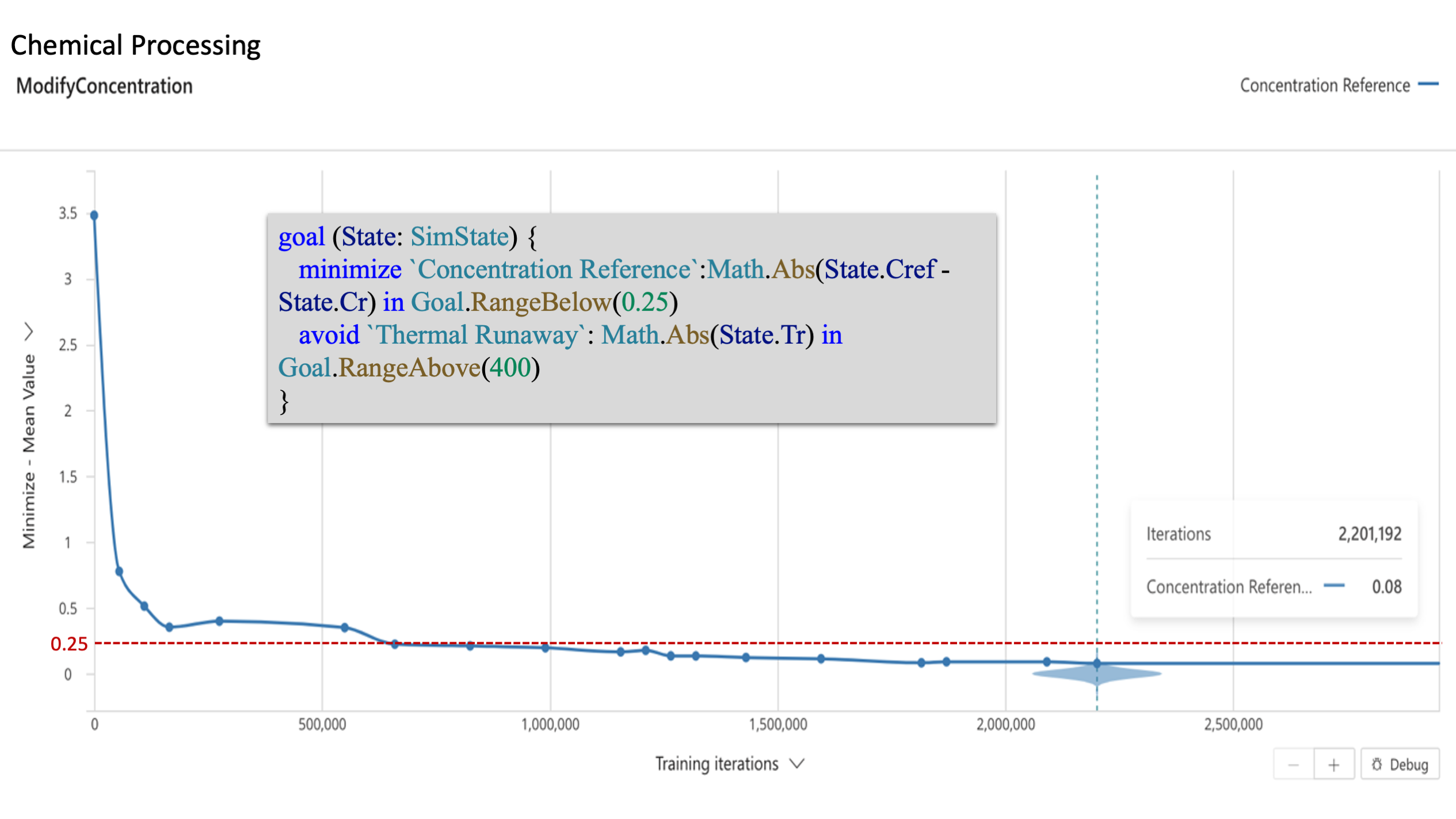}
        \caption{Minimize the difference between real Concentration level from target concentration level to below $0.25$ in a chemical reaction process}
        \label{minimizeconcentration}
    \end{subfigure}
    \caption{RL Agent Training with Goals for Optimization/Minimization Tasks}

\end{figure}

We now show the performance of the method on optimization tasks, using minimizing tasks as an example, from where we can see that even after the goals are satisfied, the policy will keep minimizing further.

We first try to minimize the distance from ball to center of the plate. We used the following specification: AVOID falling of the plate and MINIMIZE the distance from the ball to the center of the plate. For the minimize goal the minimal boundary was set as 0.01125, so the distance to target center just needs to reach 0.01125 to get to an acceptance state. Figure \ref{minimize} shows the results for this task, from where we can see that the system keeps minimizing the distance further below 0.01125 even after the goal satisfaction rate reaches 100$\%$, which is already achieved after 90,000 interactions with the environment. Figure \ref{minimizeconcentration} shows the distance of concentration level from target level for chemical reactor optimization task discussed earlier, the acceptance distance is 0.25, and we can see from the figure that the distance keeps getting even smaller than 0.25, and eventually reached 0.08.

\section{Summary}
To accelerate the adoption and applicability of RL for control and optimization tasks it is necessary to make RL easier to use and make the training process robust to as wide a range of problems as possible. However, defining a reward function is very challenging for most users, making it a key blocker for using RL.

In this work we proposed a quasi-natural-language specification language for complex control tasks, which is translated to ETLTL under the hood. We introduced a formal task specification language ETLTL, which is an extension to TLTL, to better address general optimization and control tasks that are common in industrial automation. Because previous automata representation were not suitable to represent a ETLTL formula, we also introduced a new automaton definition, SFSPA. The main difference between SFSPA and FSPA is that SFSPA does not have to terminate at the acceptance state. This allows the agent to stay in the acceptance state until an external termination criterion is met (e.g., time elapsed), allowing it to make the system increasingly more robust against disturbance.

The different elements of the proposed approach, including the additional practical implementation details described in Appendix \ref{tips}, allow for automating the training of reinforcement learning agents for real world tasks. The user can focus on problem specification instead of reward function hacking. The general applicability and ease of use of the approach was demonstrated by its successful application to a diverse set of real world problems. 

The proposed task specification language and the underlying computational framework has been implemented and deployed in the Microsoft Bonsai platform.

\section*{Acknowledgements}
The authors would like to thank Eric Traut, Victor Schnayder, Xiao Li, Ross Story, Edi Palencia, Bimal Mehta, Keen Browne, Kalyan Basu and the whole Microsoft Bonsai Team for critical feedback and insightful discussions. 

\nocite{langley00}

\bibliography{example_paper}
\bibliographystyle{icml2021}

\newpage

\appendix{{\bf \large Appendix}}

\section{Practical Implementation Details} \label{tips}
In this section, we describe implementation details that supplemented the algorithm described in the paper to make it more effective in solving real world problems.

\subsection {\bf Robustness Scaling } 
In practice, we found that bringing robustness for different variables $s$ used in predicates $f(s)<c$ to the same scale is essential for the algorithm to perform well in a general real world problems. For scaling the variables, we estimate the $min$ and $max$ for each variable by sampling values for that variable from the environment. The scale for a variable is then computed as:
\begin{equation}
    scale = max - min
\end{equation}
We divide the robustness calculated in ~\ref{robustnessCalculation} by this scale. The robustness used in practice becomes:
\begin{equation}
\rho_{scaled} = \rho / scale
\end{equation}

\subsection{\bf Robustness Boosting}
We further found that we need to do some modifications to the robustness function in section \ref{robustnessCalculation} to make the slope inside the target region steeper in order to motivate the agent to get to target region. Experimentally, we made the slope inside the target region 2 times larger than the one outside of the target region. 

\subsection {\bf Reward Capping} 
We need to cap the reward in order to prevent the agent from terminating early in fear of potential dramatically large negative reward. We cap the reward between [-MaxRobustness, MaxRobustness]. Our experiments show that MaxRobustness chosen as 2 will achieve the best result for most practical problems. 

\subsection {\bf Terminal Reward} 
When an early termination (terminating without reaching the maximum number of episode steps) occurs, a terminal reward should be assigned to the agent. Positive terminal reward should be assigned when early terminating at acceptance state, otherwise, agent may not be motivated to complete the task as early as possible. Negative terminal reward should be assigned when early terminating at trap state, otherwise, agent may go to trap state early just to avoid more negative step rewards.

The terminal reward we assign for positive terminal is:
\begin{equation}
\text{MaxRobustness}*(T - t)
\end{equation}

The terminal reward we assign for negative terminal is:
\begin{equation}
-\text{MaxRobustness}*(T - t)
\end{equation}

where, $T$ is the maximum number of steps in an episode, while $t$ is the current number of steps when the episode terminates.

\section{Goal Assessment Metrics} \label{goalassessmentmetrics}

This section lists the assessment metrics we are reporting for the current operators: Reach, Drive, Avoid, Minimize, Maximize.

\subsection{Reach} 
{\itshape reach objectiveName: testValue in targetRange } 
\begin{itemize}
\item {\bf Success}: satisfied the reach goal or not

\item {\bf SuccessRate}: percentage of episodes of the goal being successful for $N$ episodes 

\item {\bf GoalSatisfactionRate (GSR)}: a number between 0 and 1 to indicate how well this reach goal is accomplished overall 
\begin{equation}
GSR
= \text{mean}_{i=1..k}(\text{satisfactionOnDimension}_i) 
\end{equation}

Where: 

if (distanceToTarge$t_i >$ targetSiz$e_i$):\\
$\text{satisfactionOnDimension}_i$ = 0\\
else \\
$\text{satisfactionOnDimension}_i$  = \\
1 - distanceToTarge$t_i$/ targetSiz$e_i$
\end{itemize}

\subsection{Avoid} 
{\itshape avoid objectiveName: testValue in avoidRange }
\begin{itemize}
\item {\bf Success}: satisfied the avoid goal or not

\item {\bf SuccessRate}: the percentage of episodes of the goal being successful for $N$ episodes 

\item {\bf GSR}: a number between 0 and 1 to indicate how well this avoid goal is accomplished overall 
\begin{equation}
GSR = N/\text{MaxEpisodeLength}, 
\end{equation}
where $N$ is the number of iterations that agent manages to avoid the region, MaxEpisodeLength is the maximum episode length allowed in the training if the episode didn't terminate early due to other reasons. 

\end{itemize}

\subsection{Drive} 
{\itshape drive objectiveName: testValue in targetRange }
\begin{itemize}
\item {\bf Success}: satisfied the drive goal or not

\item {\bf SuccessRate}: the percentage of episodes of the goal being successful for $N$ episodes 

\item {\bf PercentageOfIterationsInTargetRegion}: the percentage of iterations the testValue is inside the target range 

\item {\bf MaxTargetReachingIterations}: the maximum number of iterations the testValue takes to reach the target range throughout the whole episode 

\item {\bf GSR}: a number between 0 and 1 to indicate how well this drive goal is accomplished overall \\
 $GSR$ = 1 – iterationsOutSideOfTargetRegion * ScaledDistance /numberIterations

Where: 

{\itshape iterationsOutSideOfTargetRegion} is the number of iterations since it has started to drive to the region once it is not in the region. 

{\itshape numberIterations} is the total number of iterations in the episode 

{\itshape ScaledDistance} is the scaled distance to target region:

   {\itshape ScaledDistance = distanceToTarget/referenceLength}

where {\itshape referenceLength} is the largest distance among the data that has been collected.

\end{itemize}

\subsection{Minimize and Maximize}
 {\itshape maximize objectiveName: testValue in targetRange }
    
 {\itshape minimize objectiveName: testValue in targetRange }
    
\begin{itemize}
\item {\bf Success}: satisfied the minimize/maximize goal or not

\item {\bf SuccessRate}:  the percentage of episodes of the goal being successful for $N$ episodes 

\item {\bf MeanValue}: the mean of the values of var in the episode 

\item {\bf Distance}: the final distance to target 

\item {\bf GSR}: a number between 0 and 1 to indicate how well this minimize/maximize goal is accomplished overall 

if reached target region: {\itshape GSR} = 1 \\
else: {\itshape GSR} = 1.0 - 
    max(0.0, min(1.0, meanDistanceToTargetRegion/targetRegionSize))

\end{itemize}

\subsection{Overall Goal Assessment Metrics: }
\begin{itemize}
\item {\bf SuccessRate}: the percentage of episodes with all the goals being successful for the past $N$ episodes 

\item {\bf OverallGoalSatisfactionRate}: the average of goal satisfaction rate of all the subgoals 

\end{itemize}

\end{document}